\documentclass[journal,a4paper,onecolumn,12pt]{ieeeconf}
\usepackage{geometry}
\def\mytitle{{FORFIS}: A forest fire firefighting simulation tool for education and research}
\def\FORFIS{{FORFIS}}
\IEEEoverridecommandlockouts
\usepackage{cite}
\usepackage[american]{babel}
\usepackage[utf8]{inputenc}
\usepackage{xcolor}
\usepackage{graphicx}
\graphicspath{ {./Pictures/} }
\usepackage{tikz}
\usepackage{subfig}
\usepackage{float}
\usepackage{url}
\usepackage{diagbox}
\usepackage{amsmath}
\usepackage{amssymb}
\newtheorem{definition}{Definition}[section]
\makeatletter
\def\endexample{\popQED\@endtheorem}
\def\term#1%
{\@nomath\term\ifdim\fontdimen\@ne\font>\z@%
\textbf{#1}\else\textit{#1}\fi}
\def\intcc#1{\ensuremath{\left[#1\right]}}
\def\intco#1{\ensuremath{\left[#1\right[}}

\def\intoo#1{\ensuremath{\left]#1\right[}}
\makeatother
\usepackage{algorithm}
\usepackage[noend]{algpseudocode}
\def\healthy{\texttt{healthy}}
\def\afire{\texttt{afire}}
\def\burnt{\texttt{burnt}}
\def\nonflam{\texttt{nonflam}}
\def\ext{\texttt{ext}}
\def\nop{\texttt{nop}}
\def\agent{\texttt{agent}}
\def\noagent{\texttt{noagent}}
\def\retardant{\texttt{retardant}}
\def\setofneighbors{B}

\author{Marvin Bredlau, Alexander Weber, Alexander Knoll 
\thanks{
The authors are with the
Munich University of Applied Sciences,
Department of Mechanical, Automotive and Aeronautical Engineering,
80335 M\"unchen, Germany.
}%
\thanks{This work has been supported by the German Federal Ministry of Education and Research  (Project ARCUS; No. 13FH734IX6) and the Government of Upper Bavaria (Project EILT; No. HAM-2109-0030). %
}}
\title{\mytitle{}}

\begin{document}
\maketitle
\thispagestyle{empty}

\begin{abstract}
We present a forest fire firefighting simulation tool 
named {\FORFIS} that is implemented in \texttt{Python}. %
Unlike other existing software, 
we focus on a user-friendly software interface 
with an easy-to-modify software engine. 
Our tool is published under GNU GPLv3 license and 
comes with a GUI as well as 
additional output functionality.
The used wildfire model is based on 
the well-established approach by cellular automata in two variants -- 
a rectangular and a hexagonal cell decomposition of the wildfire area. 
The model takes wind into account. 
In addition, our tool allows the user 
to easily include a customized 
firefighting strategy for the firefighting agents.
\end{abstract}

\section{Introduction}
\label{s:introduction}

Forest fires cause widespread damage to society. 
In addition to ecological consequences due to 
the destruction of large landscapes and 
the loss of living space, 
forest fires are a great danger 
to all living creatures in the vicinity. 
The cost for firefighting and 
repairing the destroyed areas is enormous.
Costs of more than one billion US dollars have been reported 
in some cases \cite{Diaz2012}.
Meanwhile, the changes in weather parameters, 
such as temperature and soil moisture, 
which are occurring in the course of ongoing global warming, 
are favouring the spread of forest fires. 
It is assumed that forest fires 
will become more frequent \cite{Pozniak19}.

Consequently, it is of great importance 
to optimize the strategies for fighting forest fires and 
to limit the damage. 
At the same time, the use of robotic systems 
to support or replace human operators 
is desirable to protect them 
from the hazards and 
to simplify firefighting 
even in difficult environmental conditions 
such as wind or challenging terrain.

The basis for the development or 
calculation of strategies is 
a dynamic model of a forest fire. 
There is a rich collection of works 
in the area of modelling forest fires
both with the theoretical approach and 
the practical approach 
in terms of numerical computation.
Among the earliest works on fire models 
is the one of Rothermel \cite{Rothermel72}, 
which has been cited since now more than 3000 times. 
Rothermel's approach can be 
categorized in models where the fire front 
possesses elliptical shape and travels along 
a continuous plane. 
Solutions of the dynamics of 
this approach are typically obtained by solving partial 
differential equations. 
In contrast, computationally simpler is 
the approach by \emph{cellular automata}, 
which uses time- and state-discrete dynamics. 
It was first introduced by \cite{AlbinetSearbyStauffer86} with
numerous other works following up 
\cite{ClarkeBrassRiggan94,
EncinasEncinasWhiteMartindelreySanchez07,
EncinasWhiteMartindelreySanchez07,
AlexandridisVakalisSiettosBafas08}. 
(For a survey, see \cite{Sullivan07}.)

There is a large amount of software packages 
that implement wildfire forecasting, 
with some of them being available as open-source. 
A compact summary of various tools is 
given in \cite{PapadopoulosPavlidou10}. 
The tools described therein, 
such as 
\texttt{Phoenix} \cite{TolhurstShieldsChong08}, 
the popular tool 
\texttt{FARSITE} \cite{Finney95} or 
\texttt{SPARK} \cite{MillerHiltonSullivanPrakash15} 
provide very accurate models. 
However, their downside is that they must be fed with 
a lot of location-specific data making 
a spontaneous use complicated. 
Also, the computational complexity does not allow 
to use those tools for further purposes 
like investigating firefighting strategies.
Investigating firefighting strategies on a formal level 
is a relatively new discipline, 
e.g. 
\cite{SomanathKaramanYoucef14,HaksarSchwager18,MontenegroEtAl20}, 
which is probably due to the fact that
the compute resources significantly increased in the past years.

This work introduces \texttt{\FORFIS{}}, 
which is a \underline{for}est \underline{fi}re \underline{s}imulation tool. 
The present work has been particularly motivated 
by the works \cite{HaksarSchwager18,MontenegroEtAl20}, 
which come with open-source implementations. 
Therein, a wildfire model combined with 
a deep reinforcement learning approach 
for computing a firefighting strategy are implemented.

The motivation for the development of 
our tool {\FORFIS{}} is twofold. 
Firstly, it shall attract students' attention by making 
modifications in the model parameters simple and understandable.
Secondly, due to modular structure of the software, 
it can be easily modified so that research work 
can be performed with it.
The works \cite{HaksarSchwager18, MontenegroEtAl20}
use a rectangular cellular automaton to model the wildfire dynamics. 
Our software additionally comes with a hexagonal cellular automaton, 
which leads to more accurate fire models \cite{EncinasWhiteMartindelreySanchez07}.
Our tool can be cloned from 
\begin{center}
\url{https://github.com/MBredlau/ForFiS}
\end{center}
and run using \texttt{./main.py} provided that \texttt{python3} is installed. 

We organize the remaining part of this article as follows.
In Section \ref{s:model} the implemented forest fire model 
is rigorously defined. Also,
the firefighting agents model is explained. 
Section \ref{s:gui} describes the user interfaces.

\section{Forest fire and firefighting agents model}
\label{s:model}
As already indicated, our work follows in part
the methods in \cite{HaksarSchwager18,MontenegroEtAl20}.
More specifically, we follow the approach of 
cellular automata to model wildfires
and the firefighting agents act on that model. 
The wildfire model and the agents model are discussed in 
Section \ref{ss:forestfiremodel} and Section \ref{s:agentmodel}, 
respectively.
In Section \ref{ss:prelim} 
basic notation and concepts required subsequently are introduced.
\subsection{Preliminaries}
\label{ss:prelim}
The symbols $\mathbb{N}$ and $\mathbb{R}$ stand for 
the set of positive integers and 
the set of real numbers, respectively. 
If $M$ is a set and $N \in \mathbb{N}$ then 
$M^N$ stands for the $N$-fold Cartesian product of $M$, 
e.g. $M^2 = M \times M$.
The notation $f \colon A \rightrightarrows B$ stands for 
a set-valued map with domain $A$ and 
as image subsets of $B$. 
$f$ is \term{strict} if $f(a) \neq \emptyset$ for all $a \in A$.

The next definition is crucial for the present work as it allows to 
model the forest fire and the firefighting agents in a simple and concise form.
\begin{definition}
\label{def:transitionsystem}
A \term{transition system} is a triple 
\begin{equation}
\label{e:transitionsystem}
(X,U,F)
\end{equation}
where $X$ and $U$ are non-empty sets and 
$F\colon X \times U \rightrightarrows X$ is a strict set-valued map. 
The components of \eqref{e:transitionsystem} are called \term{state space}, \term{input space} 
and \term{transition function}, respectively.
\end{definition}
A transition system \eqref{e:transitionsystem} is 
implicitly equipped with 
time-discrete dynamics of the form
\begin{equation}
\label{e:dynamics:discrete}
x(t+1) \in F(x(t),u(t)).
\end{equation}
The use of a differential inclusion 
allows to model uncertainties in the dynamics.
Thereafter, we will model a forest fire by a transition system
where the transition function is modelled by probabilities so that 
the forest fire model is basically a \term{Markov decision process}.
\subsection{Forest fire model}
\label{ss:forestfiremodel}
\begin{figure}
	\centering
	\subfloat[][]{\includegraphics[width=0.30\linewidth]{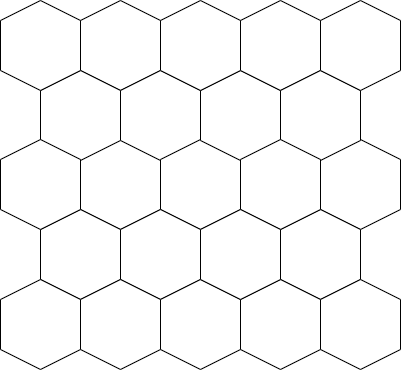}}
	\qquad
	\subfloat[][]{\includegraphics[width=0.27\linewidth]{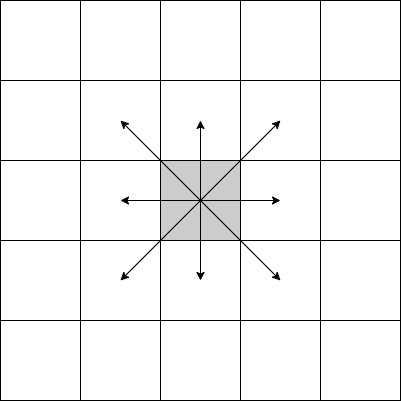}}
	\qquad
	\subfloat[][]{\includegraphics[width=0.30\linewidth]{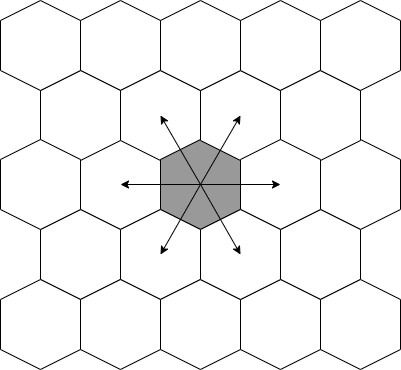}}
	\caption{\label{fig:neighbors}Illustration of the two types of partitions of the forest area implemented in \FORFIS{} and neighbouring sub-areas of a sub-area. (a) Hexagonal partition of the forest area, (b) rectangular partition; The set $\setofneighbors{}$ of unit vectors pointing to neighbouring sub-areas of the gray-coloured one are indicated. (c) Same as (b) but for the hexagonal decomposition.}
\end{figure}

To model the wildfire, 
we use a special transition system 
of the form \eqref{e:transitionsystem}.
Firefighting agents have an impact on 
the wildfire by the input of this transition system. 
In brief, we subdivide the forest area, 
which is assumed to be two-dimensional, flat and compact, 
into $N \in \mathbb{N}$ sub-areas.
Formally, we define a finite partition on the forest area and
the topological closures of the elements of 
this partition are congruent to each other.
See Fig.~\ref{fig:neighbors}(a).
An element of the partition is called \textit{sub-area}.
The particular form of the partition, 
and in particular the number of partition elements $N$, 
is a design parameter. 
Intuitively, one may think of 
a sub-area as the area occupied 
by a single tree of the forest.
The \FORFIS{} software includes two different types of partitions,
namely a rectangular and a hexagonal partition, 
cf. Fig.~\ref{fig:neighbors}. 
The integer $N$ is a user input.
Each sub-area has a set of \emph{neighbouring} sub-areas, 
which can be identified by a set of unit vectors
denoted by $\setofneighbors$. 
This set defines the direction of 
all neighbouring sub-areas to the fixed sub-area. 
For a rectangular partition, 
$\setofneighbors$ equals $\{e_1,\ldots,e_8\}$ with $e_1,\ldots,e_8 \in \mathbb{R}^2$ 
as indicated in Fig.~\ref{fig:neighbors}(b).
In the hexagonal case, 
$\setofneighbors = \{e_1,\ldots,e_6\}$ as indicated in Fig.~\ref{fig:neighbors}(c).
Each of the $N$ sub-areas possesses a discrete state and 
firefighting agents can influence this state. 
To be more precise, we define 
\begin{equation}
\label{e:statespace}
X = \{\healthy,\afire,\burnt,\ext,\nonflam\}^N,    
\end{equation}
for the transition system \eqref{e:transitionsystem} modelling the forest fire,
where $N$ is as above. 
These five discrete elements characterize a 
healthy, 
burning
or burnt sub-area, or 
a sub-area that has been extinguished.
The element \nonflam{} describes 
a sub-area that cannot burn. 
The control input is the set
\begin{equation}
\label{e:controlset}
U = \{ \texttt{nop}, \texttt{retardant}\}^N,
\end{equation}
which describes that either 
the sub-area stays untouched (element \texttt{nop} - no operation) or 
retardant has been applied (element \texttt{retardant}). 
The state of the forest fire, 
which is an element of $X$ in \eqref{e:statespace}, 
changes in discrete time according to 
the difference inclusion \eqref{e:dynamics:discrete}, 
where the transition function $F$ will be defined subsequently.
\begin{table}
    \centering
    \begin{tabular}{c|ccccc}
   \diagbox{$x(t)_i$}{$x(t+1)_i$} & \healthy & \afire & \burnt & \nonflam & \ext \\ \hline
 \healthy & $1 - p_{\texttt{ha}}(i)$ & $p_{\texttt{ha}}(i)$ & $0$ & $0$ & $0$ \\
 \afire & $0$ & $1-p_{\texttt{ab}}-p_{\texttt{ae}}(v)$ & $p_{\texttt{ab}}$  & $0$ & $p_{\texttt{ae}}(v)$ \\
  \burnt & $0$ & $0$ & $1$ & $0$ & $0$ \\
  \nonflam & $0$ & $0$ & $0$ & $1$ & $0$ \\
  \ext & $0$ & $0$ & $0$ & $0$ & $1$
\end{tabular}
\caption{Transition probabilities for the state of the $i$th sub-area at time $t$, where $v \in \{ \texttt{nop}, \texttt{retardant}\}$ and $p_{\texttt{ha}}(i)$, $p_{\texttt{ab}}$, $p_{\texttt{ae}}(v)$ are as in \eqref{e:pha}--\eqref{e:pae}.}
    \label{tab:transitionprobabilities}
\end{table}%

Firstly, we assume constant wind in the area 
specified by a vector $w \in \mathbb{R}^2$, 
which is either $0$ (``no wind'') 
or of length $1$ pointing to the direction of the wind.
Secondly, 
the subset $\setofneighbors(i)_{\afire} \subseteq \setofneighbors$ 
is defined to contain those vectors 
that point to a neighbouring sub-area of 
the $i$th sub-area that is in the states $\afire$, $i \in N$.
Thirdly, we define design parameters
\begin{equation}
\alpha_0 \in \intco{0,1}, \alpha_\textrm{wind} \in \intoo{0,1}, \ \beta \in \intcc{0,1}, \ \zeta \in \intcc{0,1},
\end{equation}
where $\alpha_0 \leq \alpha_\mathrm{wind}$. 
Intuitively, $\alpha_0$ determines the likelihood 
that a healthy area remains healthy 
in the case of no wind. 
The corresponding maximum likelihood in case of wind 
is determined by $\alpha_\textrm{wind}$.
The parameter $\beta$ controls the likelihood that 
an area on fire becomes completely burnt. 
The efficiency of retardant application is determined by $\zeta$.

\begin{figure}
	\centering
	\subfloat[][]{\includegraphics[width=0.28\linewidth]{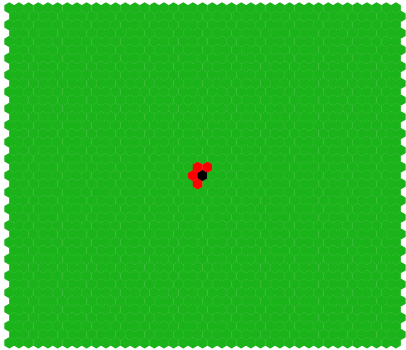}}
	\qquad
	\subfloat[][]{\includegraphics[width=0.28\linewidth]{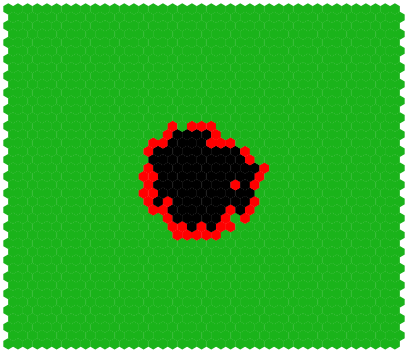}}
	\qquad
	\subfloat[][]{\includegraphics[width=0.28\linewidth]{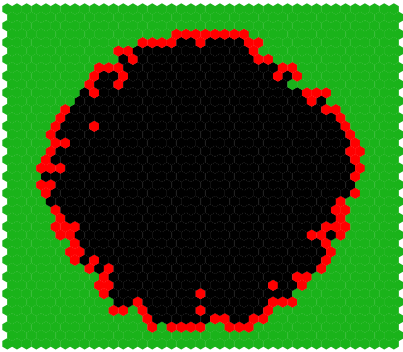}}
	\qquad
	\subfloat[][]{\includegraphics[width=0.28\linewidth]{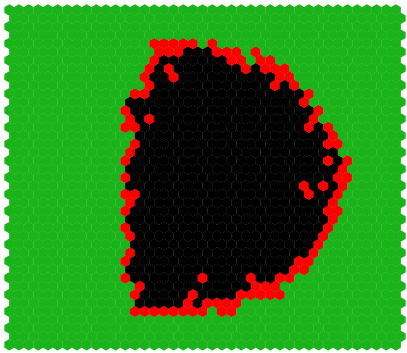}}
	\qquad
	\subfloat[][]{\includegraphics[width=0.28\linewidth]{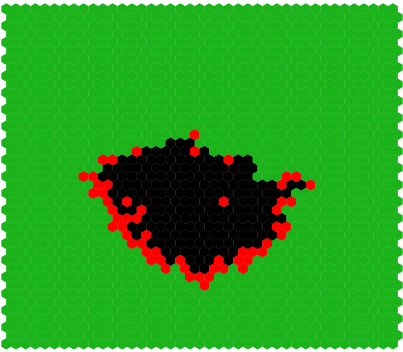}}
	\qquad
	\subfloat[][]{\includegraphics[width=0.28\linewidth]{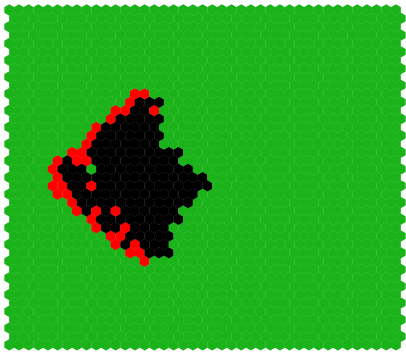}}
	\caption{\label{fig:wind}Graphical simulation output by \FORFIS{} for a hexagonal decomposition of the fire area by $42 \times 42$ hexagons and parameters $\alpha_{0}=0.7$, $\alpha_{\text{wind}}=1$, $\beta=0.6$ for $16$ time steps. (a) -- (c) Fire spread without wind. (d) -- (f) Influence of wind from the west (d), from north (e), from the east side (f) on the forest fire spread.}
\end{figure}

The transition probabilities are given by 
the following definitions in combination with 
Tab.~\ref{tab:transitionprobabilities}. 
The non-ignition likelihood depends on the wind $w$ and the
burning neighbouring area $e \in \setofneighbors$ \cite{JohnstonKelsoMilne08}:

\begin{equation}
    \alpha(e,w) = \begin{cases} \alpha_0, & \text{if } w = 0 \\
    \frac{\alpha_0 \cdot \| w \|}{1-(1-\alpha_0/\alpha_\mathrm{wind})\cdot e^\top w}, & \text{otherwise}
    \end{cases},
\end{equation}

\begin{alignat}{2}
\label{e:pha}
p_{\texttt{ha}}(i) &:= p_{\healthy,\afire}(i) && = 1 - \prod_{e \in \setofneighbors(i)_{\afire}} \alpha(e,w), \\
\label{e:pab}
p_{\texttt{ab}} &:=  p_{\afire,\burnt} && = \beta, \\
\label{e:pae}
p_{\texttt{ae}}(v) &:=  p_{\afire,\ext}(v) && = \begin{cases} \zeta, & \text{if } v = \retardant.\\
0 & \text{otherwise} \end{cases}
\end{alignat}
We note that \eqref{e:pha} reduces to $1 - \alpha_0^f$, where $f= | \setofneighbors(i)_{\afire}|$, 
in the case $w=0$. 
This expression is used in \cite{HaksarSchwager18}.
To conclude, a forest fire model is specified by the $6$-tuple
\begin{equation}
(N, \setofneighbors, \alpha_0, \alpha_\mathrm{wind}, \beta, \zeta).
\end{equation}
In Fig.~\ref{fig:wind} some graphical outputs of the \FORFIS{} software are depicted.
\paragraph*{Implementation details}
The forest fire model is implemented in the \texttt{Forest} class. 
Model configurations are made by adapting the parameters 
inside the configuration file as described later in Section \ref{s:gui}. 
The initial state $x(0) \in X$ of the fire (``fire source'') is initialized 
in the \texttt{Forest.init\_centre()} method and can be freely modified there.

The data structure for the fire model is based on a two-dimensional \texttt{numpy} array, where each cell state corresponds to an specific integer variable.
\subsection{Firefighting agents model}
\label{s:agentmodel}
The firefighting agents are also modelled by means of 
a special transition system. 
Specifically, the agents are modelled by 
the transition system $(U \times Z, X, A)$, 
where $U$ is as in \eqref{e:controlset}, 
\begin{equation*}
Z = \{ \agent, \noagent\}^N
\end{equation*}
and therefore $A \colon (U \times Z) \times X \rightrightarrows U \times Z$.
The state space of the agents transition system indicates 
for each of the $N$ sub-areas 
whether retardant has been applied (cf. \eqref{e:controlset}) and 
whether an agent is inside the sub-area (element \agent) or not (element \noagent). 
Consequently, the transition function $A$ models 
the physical movement of the agents as well as the firefighting strategy.
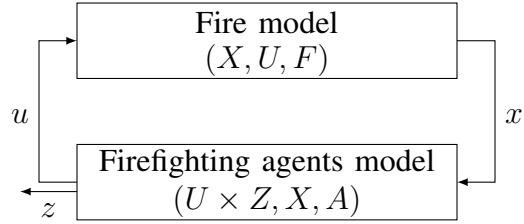
\begin{figure}
\centering
\begin{tikzpicture}
\draw  (-3.75,2.375) rectangle (1.25,1.375);
\draw  (-3.75,0.5) rectangle (1.25,-0.5);
\coordinate (v1) at (-3.75,0) {} {};
\coordinate (v2) at (-4.25,0) {} {};
\coordinate (v3) at (-3.75,1.875) {} {} {};
\draw[-latex] (v1) -- (v2) |- (v3);
\coordinate (v4) at (-3.75,-0.125) {} {} {};
\coordinate (v5) at (-4.5,-0.125) {} {} {};
\draw[-latex] (v4) -- (v5);
\coordinate (v5) at (1.25,1.875) {};
\coordinate (v6) at (1.75,1.875) {};
\coordinate (v7) at (1.25,0) {};
\draw[-latex] (v5) -- (v6) |- (v7);
\node at (-3.75/2+1.25/2,2.125) {Fire model};
\node at (-3.75/2+1.25/2,1.625) {$(X,U,F)$};
\node at (-3.75/2+1.25/2,0.25) {Firefighting agents model};
\node at (-3.75/2+1.25/2,-0.25) {$(U \times Z,X,A)$};
\node at (2,0.875) {$x$};
\node at (-4.5,0.875) {$u$};
\node at (-4.125,-0.375) {$z$};
\end{tikzpicture}
\caption{\label{fig:closedloop}Forest fire firefighting model.}
\end{figure}
The agents transition system is interconnected with the forest fire model 
as depicted in Fig.~\ref{fig:closedloop}. 
Intuitively, the agents transition system takes the state of the forest fire as its input 
and returns to the wildfire transition system an element in \eqref{e:controlset}. 
In the current version of \FORFIS, 
the agents are subject to certain restrictions as detailed below.

\paragraph{Agent's movement}
The firefighting agents can move over the whole area, 
where their position belongs exactly to one sub-area and two agents must not occupy the same sub-area.
The agent's movement is limited at every time step 
to the neighbouring cells as shown in Fig.~\ref{fig:neighbors} and 
an agent must move.

\paragraph{Agent's sensors}
Each agent is modelled as to be equipped with two sensors, 
where in the present version of the software we precisely follow \cite{HaksarSchwager18}:
\begin{itemize}
	\item[-] \textit{Radio Receiver}: All agents know the initial state of the forest fire. 
	In fact, $x(0)$ is fully available for processing. Also, the state signal of the agents transition system $(u(t),z(t))\in U \times Z$ is fully available at any time $t$.
	\item[-] \textit{Infrared Camera}: A camera facing downwards determines the states of the sub-areas located under the agent, resulting in an image of size $3 \times 3$, with the center at the agent's position. At the edges of the grid the image is padded with artificial states \nonflam. Hence, $x(t)$ is assumed not to be fully available for $t\geq 1$ but only a few components.
\end{itemize}
Moreover, every agent occupies a memory, which is a binary value initially set to false, that tracks, whether a sub-area in the state \afire{} or \burnt{} was sensed yet.

\begin{figure}
	\centering
	\subfloat[][]{\includegraphics[width=0.28\linewidth]{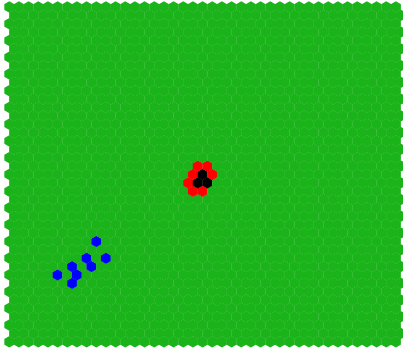}}
	\qquad
	\subfloat[][]{\includegraphics[width=0.28\linewidth]{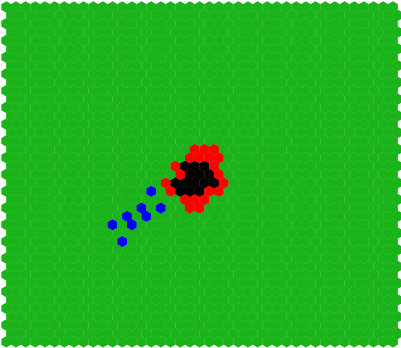}}
	\qquad
	\subfloat[][]{\includegraphics[width=0.28\linewidth]{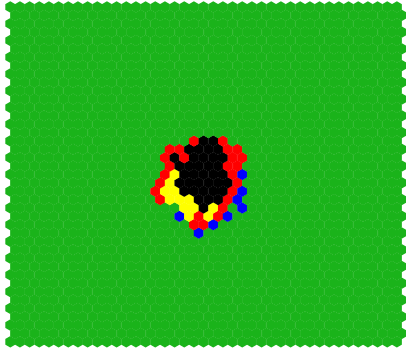}}
	\qquad
	\subfloat[][]{\includegraphics[width=0.28\linewidth]{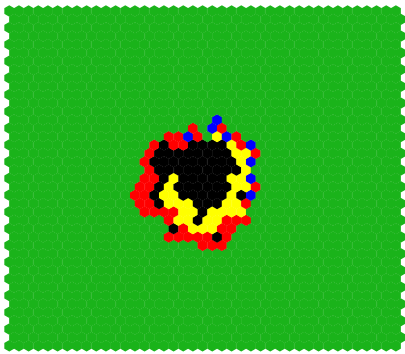}}
	\qquad
	\subfloat[][]{\includegraphics[width=0.28\linewidth]{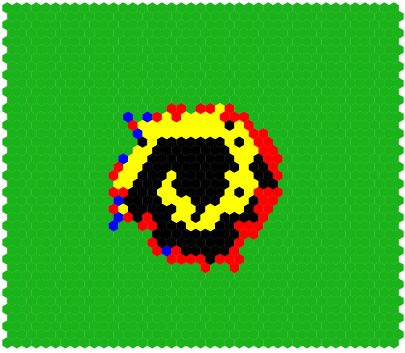}}
	\qquad
	\subfloat[][]{\includegraphics[width=0.28\linewidth]{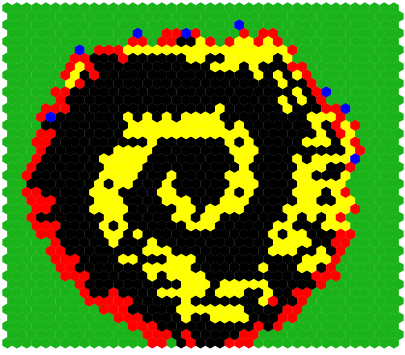}}
	\caption{\label{fig:Heuristik}Graphical simulation output of \FORFIS{} for $6$ selected time steps with $8$ firefighting agents following the heuristic strategy in \cite{HaksarSchwager18}. 
	The parameters are $42 \times 42$ hexagons for the partition, 
	$\alpha_0=0.5$, 
	$\beta=0.6$, $\zeta=1$.}
\end{figure}

In the current version of \FORFIS, 
a heuristic firefighting strategy following \cite{HaksarSchwager18}
is implemented. The firefighting process is divided into two phases. In the first phase, all agents move on a direct path towards the initial fire source in order to arrive at the fire as quickly as possible.
The second phase is initiated when the agent's memory is set to true. In this phase, the agents move counterclockwise along the fire front. Whenever an agent perceives a tree in state \afire{} below it, retardant is applied.
A formal definition of the transition function $A$ is omitted in this note
due to its complexity. 
The influence of firefighting agents
following this heuristic is illustrated in Fig.~\ref{fig:Heuristik}.

\begin{figure}
	\centering
    \includegraphics[width=\linewidth]{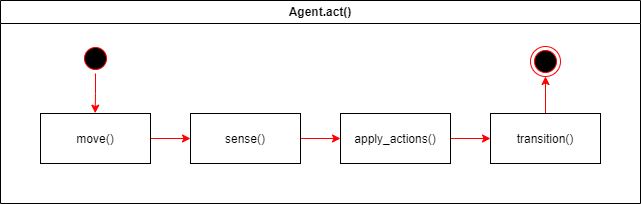}
	\caption{Sequence of the agent actions in the method \texttt{agent.act()}}
	\label{fig:agent:act}
\end{figure}

\paragraph{Implementation details}
The agents act with a sequence of three methods, 
which are called within the \texttt{agent.act()} method 
as part of the \texttt{agent} class (refer to Fig.~\ref{fig:agent:act}). 
In \texttt{move()}, 
the position of each agent is updated 
according to the strategy. 
\texttt{Sense()} returns the sensor data and 
\texttt{apply\_actions()} applies the control actions 
according to the strategy. 
The transitions are divided into two parts. 
After every call of \texttt{agent.act()} 
the control actions are considered for a transition step. 
After six steps the whole forest updates without the agent impact.
For the sake of including new firefighting strategy algorithms 
there is a transparent and easy interface included in \FORFIS. 
As part of the \texttt{agent} class, 
the method \texttt{act()} 
applies the actions as described in Section \ref{s:agentmodel}.

The strategy consists of two parts: 
Trajectory and control actions. 
The trajectory is executed in the \texttt{move()} method and 
the control actions in the \texttt{apply\_actions()} method. 
The algorithm to calculate the strategy can be chosen freely 
in those methods as long as the following interfaces are retained:
\begin{itemize}
    \item \texttt{move()} gets the unique identifier and the position of an agent as the input and returns the new position.
    \item \texttt{sense()} sets the sensors and the memory with the described properties from Section \ref{s:agentmodel}.
    \item \texttt{apply\_actions()} takes the unique identifier of the agent and its position as the input and returns the boolean control action value with $0$ for $\nop$ and $1$ for $\retardant$.
\end{itemize}

For the user to implement its own algorithm, 
set the parameter \textit{strategy} in the 
\texttt{config.yaml} file to \textit{user} and 
use the template methods in the separated file named \texttt{user\_strategy.py}.
\subsection{Firefighting cost function}
To evaluate the success of the firefighting strategy 
the user can define a metric 
by adapting the method \texttt{success\_metric()} 
as part of the \texttt{Forest} class. 
The method calls \texttt{calc\_statistic()}, 
which returns the number of sub-areas in the states \healthy, \afire, \burnt{} and \ext. 
The \texttt{success\_metric()} method shall return the calculated cost defined by this metric.

The default cost function is 
\begin{equation}
    \label{e:cost}
    G(x(t),t) = c_{\healthy} \cdot N_\healthy(x(t))+c_{\ext} \cdot N_\ext(x(t))+c_{\textrm{time}} \cdot t,
\end{equation}
where
$N_q(p) = | \{ i \in \{1,\ldots,N\} \mid p_i = q\} |$, i.e. $N_q(p)$ is the 
number of sub-areas being in the state $q$.
The coefficients $c_{\healthy}$, $c_{\ext}$ and $c_{\textrm{time}}$ are specified in the \texttt{config.yaml} file.
\section{Graphical user interface and configuration file}
\label{s:gui}
\begin{figure}
	\centering
	\includegraphics[width=0.6\textwidth]{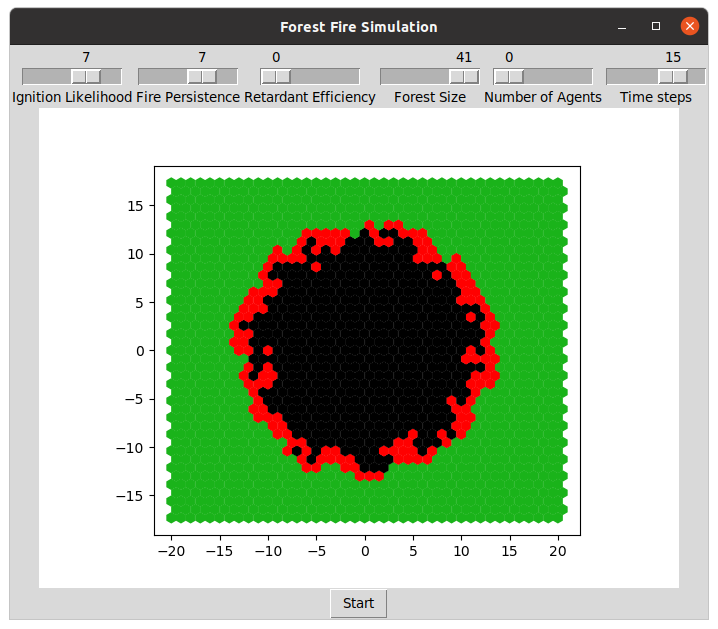}
	\caption{\label{fig:GUI}Graphical user interface of FORFIS}
\end{figure}
The graphical user interface (see Fig.~\ref{fig:GUI}) 
provides a button to start a new simulation, 
a visualization area and 
the following basic model-specific and general modification parameters:
\begin{itemize}
    \item \textit{Ignition Likelihood} $l$: $l \in [0,10]$ such that $\alpha = 1-\frac{l}{10}$
    \item \textit{Fire Persistence} $p$: $p \in [0,10]$ such that $\beta = \frac{p}{10}$
    \item \textit{Retardant Efficiency} $e$: $e \in [0,10]$ such that $\zeta = \frac{e}{10}$
    \item \textit{Forest Size} $n$: Creates a grid of $n\times n$ cells
    \item \textit{Number of Agents} $a$: Includes $a$ agents. $0$ for a forest fire simulation without firefighting agents.
    \item \textit{Time Steps} $\tau$: Simulation will last $\tau$ time steps. Choose $0$ for a full simulation until the fire is extinguished or fully burnt.
\end{itemize}
The described configurations as well as the following additional options 
can also be predefined in a configuration file called \texttt{config.yaml}:
\begin{itemize}
    \item \textit{GUI}: Boolean value to run the simulation with or without the graphical user interface
    \item \textit{agent\_mode}: Strategy algorithm, $Haksar$ for heuristic inspired by \cite{HaksarSchwager18} or $user$ for user-defined algorithm. See Section \ref{s:agentmodel}.
    \item \textit{logfile}: Boolean value to save the simulation results in a log file named with the start date and time of the simulation.
    \item \textit{grid}: \textit{rectangular} for a rectangular cell geometry, \textit{hexagonal} for a hexagonal cell geometry. 
\end{itemize}
\bibliography{mrabbrev.bib,lit.bib}
\end{document}